\pdfoutput=1

\documentclass[11pt]{article}
\usepackage{acl}

\usepackage{times}
\usepackage{latexsym}

\usepackage[T1]{fontenc}

\usepackage[utf8]{inputenc}
\usepackage{kotex}
\usepackage{microtype}
\usepackage{booktabs}
\usepackage{amsmath}
\usepackage{flexisym}
\usepackage{dcolumn}
\usepackage{kotex}
\usepackage{hhline}
\usepackage{verbatim}
\usepackage{multirow}
\usepackage{amssymb}
\usepackage{rotating}
\usepackage{subfigure}
\usepackage{dblfloatfix}
\usepackage{enumitem}
\usepackage{framed}
\usepackage[utf8]{inputenc}
\usepackage[english]{babel}
\usepackage{xcolor}
\usepackage{tabulary}
\usepackage{color}
\usepackage{arydshln}
\usepackage{ulem}
\usepackage{soul,color}

\usepackage[symbol]{footmisc}

\renewcommand*{\thefootnote}{\fnsymbol{footnote}}
\newcolumntype{L}[1]{>{\raggedright\let\newline\\\arraybackslash\hspace{0pt}}m{#1}}
\newcolumntype{C}[1]{>{\centering\let\newline\\\arraybackslash\hspace{0pt}}m{#1}}
\newcolumntype{R}[1]{>{\raggedleft\let\newline\\\arraybackslash\hspace{0pt}}m{#1}}

\title{Masked Summarization to Generate Factually Inconsistent Summaries\\ for Improved Factual Consistency Checking}
\author{Hwanhee Lee$^{1*}$, Kang Min Yoo$^{2}$, Joonsuk Park$^{2,3}$, Hwaran Lee$^{2\dagger}$ and Kyomin Jung$^{1\dagger}$ \\
  $^{1}$Seoul National University,
  $^{2}$NAVER AI Lab,
  $^{3}$University of Richmond\\
  \texttt{\{wanted1007,kjung\}@snu.ac.kr}\\
  \texttt{\{kangmin.yoo, hwaran.lee\}@navercorp.com}\\
  \texttt{park@joonsuk.org}
  }

\begin{document}
\maketitle

\footnotetext{\textsuperscript{*}Work done during an internship at NAVER AI Lab.}
\footnotetext{\textsuperscript{$\dagger$}Corresponding authors.}

\renewcommand*{\thefootnote}{\arabic{footnote}}
\setcounter{footnote}{0}

\begin{abstract}
Despite the recent advances in abstractive summarization systems, it is still difficult to determine whether a generated summary is factual consistent with the source text. To this end, the latest approach is to train a factual consistency classifier on factually consistent and inconsistent summaries. Luckily, the former is readily available as reference summaries in existing summarization datasets. However, generating the latter remains a challenge, as they need to be factually inconsistent, yet closely relevant to the source text to be effective. In this paper, we propose to generate factually inconsistent summaries using source texts and reference summaries with key information masked. Experiments on seven benchmark datasets demonstrate that factual consistency classifiers trained on summaries generated using our method generally outperform existing models and show a competitive correlation with human judgments. We also analyze the characteristics of the summaries generated using our method. We will release the pre-trained model and the code at \url{https://github.com/hwanheelee1993/MFMA}.

\end{abstract}

\section{Introduction}
\begin{figure}[!t]
\small
\begin{framed}
\textbf{Article:} \hl{Guus Hiddink}, \hl{the Russia and Chelsea coach}, has had much to smile about in his 22-year managerial career. ,…, Enjoying \hl{success} around \hl{the world} -- at \hl{different levels} with different players in \hl{different cultures} -- has made \hl{Guus Hiddink} one of the most admired bosses around. ,…, Hiddink's resume includes \hl{stints in other high-pressure jobs} such as Fenerbahce, Valencia and Real Madrid. ,…, But the \hl{straight-speaking Dutchman} is loyal to the project he has in charge of the Russian national side and insists he will leave \hl{Chelsea} at the \hl{end of the season} regardless.\\

\textbf{Reference Summary:} Born in 1946, \hl{Hiddink} has become one of \hl{the best managers} in the world . Dutchman has enjoyed \hl{huge success} at \hl{club and international level}. He's currently \hl{coach} of Russia and is in charge of \hl{Chelsea} until \hl{end of the season}.\\

\textbf{Mask-and-fill Summary Without Article:} \\ Born in 1946, \textcolor{blue}{\textit{Dutchman}} has become one of \underline{\textcolor{red}{\textit{the most respected politicians}}} in the world. Dutchman is enjoyed \underline{\textcolor{red}{\textit{success at the Olympics and World Cup}}}. He's currently the \underline{\textcolor{red}{\textit{President of Russia}}} and is in charge of \underline{\textcolor{red}{\textit{the country}}} until the end of the season.\\

\textbf{Mask-and-fill Summary With Masked Article:} \\ Born in 1946, \textcolor{blue}{\textit{Hiddink}} has become one of \textcolor{blue}{\textit{the most admired managers}} in the world. Dutchman has enjoyed \textcolor{blue}{\textit{successful spells}} at \underline{\textcolor{red}{\textit{Chelsea and Real Madrid}}}. He's currently \textcolor{blue}{\textit{manager of Russia}} and is in charge of \underline{\textcolor{red}{\textit{the country}}} until the end of the season.

\end{framed}
\caption{
An example of generated negative summary using masked article. Spans that are highlighted are masked when generating the negative summary. Note that red spans are factually inconsistent with the given article and blue spans are factually consistent.
} 
\vspace{-5mm}
\label{fig_intro}
\end{figure}



As textual content available on- and offline explodes, automated text summarization is becoming increasingly crucial~\cite{el2020automatic}; with the advances in neural text generation methods, abstractive summarization systems that generate paraphrases are quickly replacing extractive ones that simply select essential sentences from the source text~\cite{nallapati2017summarunner}. While abstractive summaries can be more coherent and informative (given the same length) than their extractive counterparts, they frequently contain information inconsistent with the source text. This is a critical issue, as it directly affects the reliability of the generated summaries.~\cite{cao2018faithful, zhao-etal-2020-reducing, maynez2020faithfulness}. 


Unfortunately, existing approaches to identify such factual inconsistency without constructing new resources have not been satisfactory. Directly measured similarity between the summary and its source text---using popular n-gram similarity metrics such as ROUGE~\cite{lin-2004-rouge} and BLEU~\cite{papineni-etal-2002-bleu}---exhibits low correlation with human judgments for factual consistency. Also, leveraging related tasks---such as natural language inference (NLI)~\cite{bowman2015large} and fact verification~\cite{thorne-etal-2018-fact}---is not ideal. This is because these tasks aim to identify relations between two sentences, whereas factual consistency checking involves a multi-sentence summary and an even longer source text~\cite{bora2018natural, falke2019ranking}.


A remaining solution is to train a factual consistency classifier with a dataset specifically constructed for this purpose. Note that \textit{\textbf{positive summaries}} are readily available. That is, the reference summaries from existing text summarization datasets can be assumed to be factually consistent with the respective source texts. Thus, the main challenge is in generating effective \textit{\textbf{negative summaries}}, i.e., summaries that are factually inconsistent with the source text. Recent works generate negative summaries by simply replacing keywords in the reference summaries or sentences extracted from the source texts~\cite{kryscinski-etal-2020-evaluating, yin-etal-2021-docnli}. 
This, however, results in negative summaries that significantly diverge from the source texts and positive summaries, which is not ideal for training factual consistency classifiers. For instance, Figure~\ref{fig_intro} shows that \textit{coach} in the reference summary is changed to \textit{President of Russia}, which is an inconsistency that is too obvious.


In this paper, we propose a novel method, Masked-and-Fill with Masked Article (MFMA),
where parts of the source text and reference summary are masked and later inferred to generate a plausible but factually inconsistent summary.
Experiments on seven benchmark datasets demonstrate that factual consistency classifiers trained on negative summaries generated with our method mostly outperform existing models and show a competitive correlation with human judgment.
We also analyze the characteristics of the negative summaries generated.
Our main contributions are as follows:
\begin{itemize}[noitemsep,leftmargin=*]
\item We propose a novel negative summary generation method for training factual consistency classifiers for abstractive summaries.
\item We show the efficacy of our method on seven benchmark datasets using classification performance and correlation with human judgment.
\item We analyze the characteristics, such as affinity and diversity, of the negative summaries generated using our method.

\end{itemize}

\begin{figure*}[h]
\small
\centering
\includegraphics[width=2.0\columnwidth]{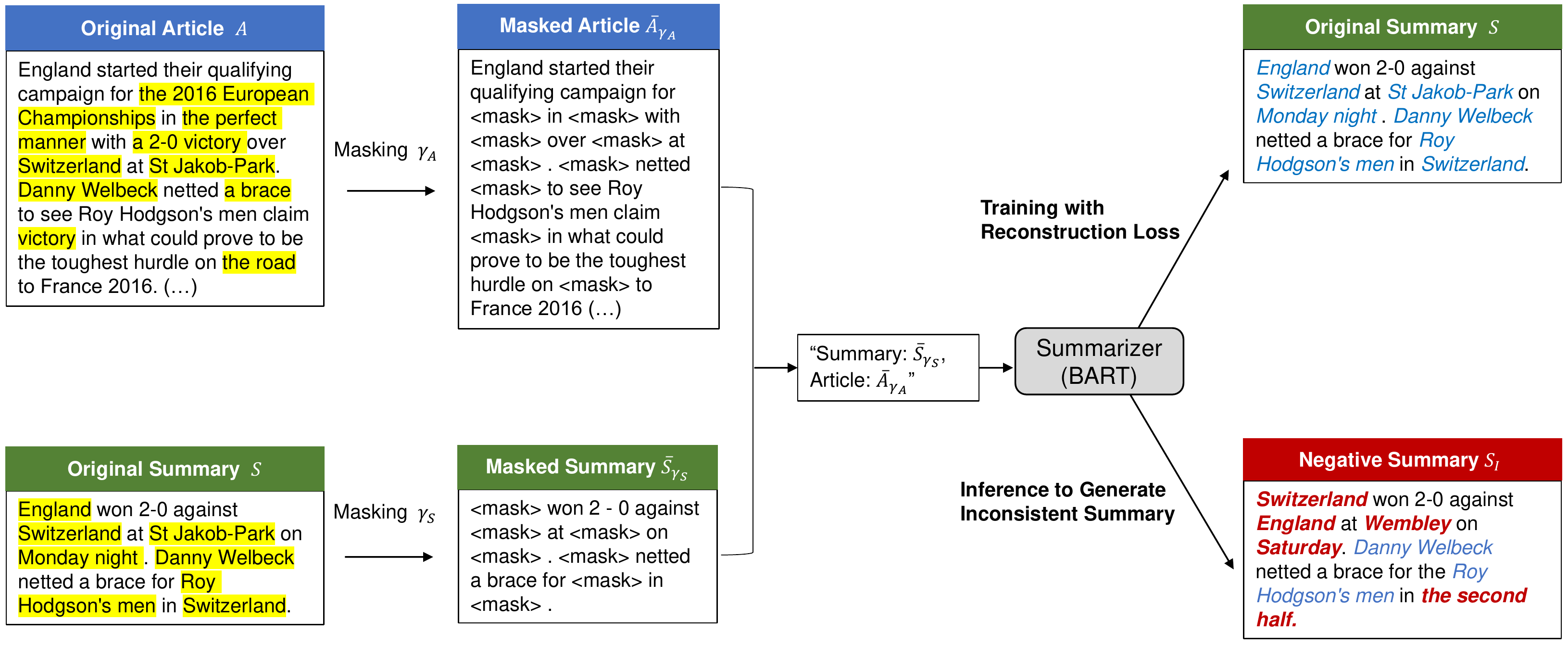}
\caption{
Overall flow of our proposed negative summary generation method Mask-and-Fill-with-Masked Article.}
\label{fig_overall}
\vspace{-3mm}
\end{figure*} 

\section{Related Work}

\subsection{Factual Inconsistency in Summarization Systems}
Previous works~\cite{maynez2020faithfulness,zhao-etal-2020-reducing, cao2018faithful} have studied the factual inconsistency in abstractive summarization systems. Especially, \cite{cao2018faithful} demonstrates that 30\% of the model generated summaries have at least one factual error, and this obstacle the practical usage. \cite{maynez2020faithfulness} specifies these factual errors in the abstractive summarization system into two types: \textit{intrinsic errors} and \textit{extrinsic errors}. Intrinsic errors occur using the contents present
in the source article like "Switzerland" and "England" in the negative summary example in Figure~\ref{fig_overall}. On the other hand, extrinsic errors are the errors generated by ignoring the source article when generating summaries. "in the second half" in Figure~\ref{fig_overall}, which is not included in the source article, is an example of extrinsic errors.

In this work, we propose a system for detecting these various factual errors that are necessary for developing a summarization system. We propose a unified method for intentionally modeling both types of errors to build a dataset for training this system. 
\subsection{Measuring Factual Consistency}
As a better way to evaluate the factual consistency, recent works such as QAGS~\cite{wang-etal-2020-asking} and QuestEval~\cite{scialom-etal-2021-questeval} adopt question generation and question answering frameworks to evaluate the factual consistency. Both methods firstly generate questions using entities or noun phrases in the candidate summary and then compare the answers of these questions between the source and the summary. Although these methods do not require any reference summaries, they have a higher correlation with human judgments than previous metrics in consistency checking. Also, the generated questions and their answers are easily interpretable. But due to their complicated structure, the computational complexity of these methods is relatively heavy and the errors in each component can be cascaded.

Following the idea that all of the contents in the summaries should be entailed by source document, models from the related tasks such as Natural Language Inference(NLI)~\cite{bowman2015large, williams2018broad, falke2019ranking} are also used to verify the factual consistency of the summaries. These approaches are simpler and more intuitive than QA-based metrics. But the data pairs in these datasets are usually composed of single sentences, and this makes it difficult to be directly used for factual consistency checking in summarization where the task requires multi-sentence level reasoning. For this reason, two recent studies FactCC~\cite{kryscinski-etal-2020-evaluating} and DocNLI~\cite{yin-etal-2021-docnli} have studied ways to make synthetic datasets for training factual consistency checking model. Both works create synthetic negative summaries using the pre-defined rules such as entity substitution or mask-and-fill. In this paper, we propose a more general negative summary generation method additionally using the masked source.\\
CoCo~\cite{xie2021factual} compares the likelihood of the generated summaries using the original source and the masked source to estimate the counterfactual samples. Different from CoCo, our work directly augments the negative summaries and train the classifier using them.

\section{Methods}

For a given article $A$ and a summary $S$, we aim to develop a factual consistency checking system that can evaluate whether $S$ is factual consistent with $A$. In other words, the system is required to discriminate a factual consistent summary $S_C$ with the factual inconsistent summary $S_I$ that consists of at least one factual error. We consider this problem as a classification task between $S_C$ and $S_I$. However, large-scale human-annotated training datasets for this task have not been constructed yet, especially for the inconsistent summaries $S_I$.

In this paper, we focus on effective augmentation methods of the inconsistent summaries. In order for that, there are two crucial conditions: 1) guarantee of inconsistency; the generated summaries should be indeed inconsistent with the source article, 2) relevance to the source article; the generated summaries should include contents related to the article. These two factors are in trade-off relations, which means that when the generated summaries are strongly inconsistent they might not be related to the article and vice versa. Therefore appropriate negative summary augmentation is required to improve the factual consistency classifier.

To generate confusing and hard negative summaries, we propose a summary generation using a masked article and a masked reference summary where some salient information is hidden.
By doing so, we let the summarizer model infer hidden information through the masked article to generate plausible negative summaries.
Note that, previous works such as FactCC and DocNLI generate negative summaries $S_I$ by changing positive summaries $S_C$ through entity replacements or mask-and-fill methods without referring to the source article.
We observe that previous methods can easily guarantee negativeness, but they often generate summaries that are very irrelevant to the source article or unnatural as shown in Figure~\ref{fig_intro}.

\subsection{Mask-and-Fill with Masked Article}




To model inconsistent summaries but related to the article, we propose a method, \textbf{M}ask-and-\textbf{F}ill with \textbf{M}asked \textbf{A}rticle (MFMA), which generates negative summaries with masked articles and masked reference summaries, as shown in Figure ~\ref{fig_overall}.

Specifically, we assumed \textit{noun phrases} and \textit{entities} in the articles are salient information, and mask them with the ratio of $\gamma_A$, resulting in masked article $\overline{A}_{\gamma_A}$.
Similarly, we also mask the salient spans in the positive summary, i.e., reference summary, with the ratio of $\gamma_S$ to form a masked summary $\overline{S}_{\gamma_S}$. Then, we concatenate $\overline{A}_{\gamma_A}$ and $\overline{S}_{\gamma_S}$ by prepending prefix token for each input text (i.e., “Summary: $\overline{S}_{\gamma_S}$, Article:  $\overline{A}_{\gamma_A}$") as shown in Figure~\ref{fig_overall}.
Next, we train a summarizer based on an encoder-decoder model, BART~\cite{lewis2020bart}, to reconstruct the original summary $S$ with the following loss:

\begin{align}
\mathcal{L} &= \sum_t -\log P({S_t | S_{<t}, [\overline{S}_{\gamma_S}; \overline{A}_{\gamma_A}]}).
\label{eq_1}
\end{align}

After training, we generate negative summaries of unseen and masked article-summary pairs through inference.
Obviously, if the mask ratio is high enough, the model is hard to correctly fill the masked contents from the erased article and reference summary. 
However, we assume the trained reconstruction model is able to fill the masks with plausible contents by inferring the related contents with the masked article.

\subsection{Masked Summarization}


As a variant of MFMA, we also study another negative summary generation model, \textbf{M}asked \textbf{S}u\textbf{M}marization(MSM). The model aims to generate summaries using masked articles $\overline{A}_{\gamma_A}$ but without masked reference summaries as follows:

\begin{align}
\mathcal{L} &= \sum_t -\log P({S_t | S_{<t}, \overline{A}_{\gamma_A}}).
\label{eq_1}
\end{align}

\noindent The MSM model is trained to generate the entire summaries without the information guidance of masked reference summaries, so MSM has merits in generating more diverse summaries than MFMA. 


\subsection{Training Factual Consistency Checking Model}
Finally, for the factual consistency checking model, we train a binary classifier of consistent summaries and inconsistent generated summaries. The pair of summary and the corresponding article are concatenated and then fed into the classification model as an input. We fine-tuned the pre-trained \mbox{ELECTRA}~\cite{clark2019electra} by adding a classifier head with binary cross-entropy loss.

\begin{table*}[h]
\small
\centering

\caption{Macro F1-score(F1) and class-balanced accuracy(BA) of the human annotated factual consistency for the benchmark datasets based on CNN/DM.}

\begin{tabular}
{L{0.2\columnwidth}C{0.11\columnwidth}C{0.11\columnwidth}C{0.11\columnwidth}C{0.11\columnwidth}C{0.14\columnwidth}C{0.14\columnwidth}C{0.16\columnwidth}C{0.16\columnwidth}C{0.11\columnwidth}C{0.11\columnwidth}}

\toprule
           \multicolumn{1}{c}{\textbf{Dataset}} & \multicolumn{2}{c}{\textbf{FactCC-Test}} & \multicolumn{2}{c}{\textbf{SummEval}} & \multicolumn{2}{c}{\textbf{QAGS-CNN/DM}} & \multicolumn{2}{c}{\textbf{FRANK-CNN/DM}} & \multicolumn{2}{c}{\textbf{Average}}\\
\cmidrule{1-11}
\multicolumn{1}{c}{\textbf{Metric}} & \textbf{F1} & \textbf{BA} & \textbf{F1} & \textbf{BA} & \textbf{F1} & \textbf{BA} & \textbf{F1} & \textbf{BA} & \textbf{F1} & \textbf{BA} \\
\midrule
\textit{\textbf{Baselines}} &  &  &  &  &  &  &  &  &  &\\
\textbf{FactCC} & 71.0 & 71.3 &	65.1 & 68.2 & 69.3 & 69.6 &	64.1 & 63.9 & 67.4 & 68.2 \\
\textbf{DocNLI} & 67.2 & 71.0 &	\textbf{71.5} & \textbf{71.3} & 62.4 & 66.2 &	66.0 & 66.0 & 66.8 & 68.6\\
\textbf{MNLI} & 55.0 & 56.0 & 51.7 & 51.7 &	48.6 & 53.4 & 50.4 & 53.3 & 51.4 & 53.6\\
\textbf{FEVER} & 57.9 & 56.2 & 52.6 & 53.6 & 39.4 & 53.3 & 49.8 & 55.6 & 49.9 & 54.7\\
\textbf{MF} & 59.9 & 64.1 &	68.2 & 67.5 & 47.6 & 56.9 & 62.4 & 62.7 & 59.5 & 62.8\\
\midrule
\textit{\textbf{Ours}} &  &  &  &  &  &  &  &  &  &\\
\textbf{MFMA} & \textbf{79.7} & \textbf{84.5} & 71.3 & 69.6 & \textbf{70.5} & \textbf{72.3} & \textbf{69.5} & 69.2 & \textbf{72.8} & \textbf{73.9}\\
\textbf{MSM} & 70.6 & 72.7 & 66.8 & 68.2 & 67.6 & 68.7 & 69.6 & \textbf{69.3} & 68.6 & 69.7\\
\bottomrule 
\end{tabular}

\label{table_accuracy_cnndm}
\end{table*}

\begin{table*}[h]
\small
\centering

\caption{Macro F1-score(F1) and class-balanced accuracy(BA) of the human annotated factual consistency for the benchmark datasets based on XSum.}

\begin{tabular}
{L{0.17\columnwidth}C{0.12\columnwidth}C{0.12\columnwidth}C{0.12\columnwidth}C{0.12\columnwidth}C{0.14\columnwidth}C{0.14\columnwidth}C{0.11\columnwidth}C{0.11\columnwidth}}
\toprule
\cmidrule{1-9}
           \multicolumn{1}{c}{\textbf{Dataset}} & \multicolumn{2}{c}{\textbf{XSumHall}} & \multicolumn{2}{c}{\textbf{QAGS-XSum}} & \multicolumn{2}{c}{\textbf{FRANK-XSum}} & \multicolumn{2}{c}{\textbf{Average}}\\
\midrule
\multicolumn{1}{c}{\textbf{Metric}} & \textbf{F1} & \textbf{BA} & \textbf{F1} & \textbf{BA} & \textbf{F1} & \textbf{BA} & \textbf{F1} & \textbf{BA} \\
\midrule
\textit{\textbf{Baselines}} &  &  &  &  &  &  &\\
\textbf{FactCC} & 52.1 & \textbf{61.8} & 63.6 & 63.7 & 50.7 & 58.0 & 55.5 & 61.2\\
\textbf{DocNLI} & 55.1 & \textbf{56.4} & 65.3 & 66.0 & \textbf{60.3} & \textbf{63.4} & 60.2 & \textbf{61.9}\\
\textbf{MNLI} & 33.3 & 52.1 & 45.2 & 51.1 &	28.8 & 50.6 & 35.8 & 51.3\\
\textbf{FEVER} & 53.1 & 55.5 & 62.2 & 63.7 & 54.9 & 63.5 & 56.7 & 60.9 \\
\textbf{MF} & 53.6 & 53.3 &	54.6 & 54.9 & 55.7 & 55.3 & 54.6 & 54.5 \\
\midrule
\textit{\textbf{Ours}} &  &  &  & \\
\textbf{MFMA} & \textbf{55.5} & 56.0 & \textbf{66.6} & \textbf{67.0} & 59.6 & 59.6 & \textbf{60.6} & 60.9\\
\textbf{MSM} & 52.6 & 53.9 & 50.8 & 55.5 & 50.8 & 51.3  & 51.4 & 53.6\\

\bottomrule 
\end{tabular}

\label{table_accuracy_xsum}
\end{table*}

\begin{table*}[h]
\small
\centering

\caption{Summary level Pearson Correlation($r$) and Spearman's Correlation($\rho$) between various automatic metrics and human judgments of factual consistency for the model generated summaries. Note that we use the confidence of consistency label for entailment based metrics.}

\begin{tabular}
{L{0.23\columnwidth}C{0.12\columnwidth}C{0.12\columnwidth}C{0.12\columnwidth}C{0.12\columnwidth}
C{0.12\columnwidth}C{0.12\columnwidth}C{0.12\columnwidth}C{0.12\columnwidth}C{0.12\columnwidth}C{0.12\columnwidth}C{0.12\columnwidth}C{0.12\columnwidth}}
\toprule

\multicolumn{1}{c}{\textbf{Dataset}} & \multicolumn{2}{c}{\textbf{SummEval}} & 
\multicolumn{2}{c}{\textbf{QAGS-CNN/DM}} & \multicolumn{2}{c}{\textbf{QAGS-XSum}} &
\multicolumn{2}{c}{\textbf{FRANK-CNN/DM}} & \multicolumn{2}{c}{\textbf{FRANK-XSum}}\\

\cmidrule{1-11}
 \textbf{Metric}
              & \textbf{$r$} & \textbf{$\rho$} 
              & \textbf{$r$} & \textbf{$\rho$}
              & \textbf{$r$} & \textbf{$\rho$}
              & \textbf{$r$} & \textbf{$\rho$}
              & \textbf{$r$} & \textbf{$\rho$}\\
\midrule
\textit{\textbf{Baselines}} &  &  &  & \\
\textbf{ROUGE-L} & 0.16 & 0.14 & 0.29 & 0.24 & 0.13 & 0.13 & 0.16 & 0.13 & 0.16 & 0.13 \\
\textbf{BLEU-4} & 0.11 & 0.12 & 0.18 & 0.23 & 0.03 & 0.03 & 0.16 & 0.17 & 0.11 & 0.14 \\
\textbf{METEOR} & 0.18 & 0.16 & 0.26 & 0.25 & 0.11 & 0.12 & 0.29 & 0.28 & 0.18 & 0.16 \\
\textbf{BERTScore} & 0.16 & 0.14 & 0.37 & 0.36 & 0.11 & 0.13 & 0.33 & 0.30 & 0.19 & 0.17\\
\midrule
\textbf{QuestEval} & 0.35 & 0.30 & 0.42 & 0.36 & 0.20 & 0.20 & 0.46 & 0.41 & 0.19 & 0.18\\
\textbf{CoCo} & 0.42 & 0.36 & \textbf{0.67} & 0.57 & 0.20 & 0.18 & 0.50 & 0.45 & 0.14 & 0.12 \\
\textbf{FactCC}  & 0.38 & 0.36 & 0.45 & 0.48 & 0.30 & 0.30 & 0.32 & 0.36 & 0.09 & 0.08 \\
\textbf{DocNLI}  & 0.51 & \textbf{0.41} & 0.60 & 0.59 & 0.36 & 0.35 & 0.49 & \textbf{0.49} & \textbf{0.25} & \textbf{0.21} \\
\textbf{MNLI}  & 0.11 & 0.13 & 0.19 & 0.22 & 0.08 & 0.10 & 0.15 & 0.16 & 0.02 & 0.03 \\
\textbf{FEVER}  & 0.33 & 0.32 & 0.40 & 0.34 & \textbf{0.38} & \textbf{0.41} & 0.38 & 0.43 & 0.20 & 0.19 \\
\textbf{MF}  & 0.44 & 0.35 & 0.43 & 0.30 & 0.10 & 0.10 & 0.40 & 0.39 & 0.10 & 0.13 \\
\midrule
\textit{\textbf{Ours}} &  &  &  & \\
\textbf{MFMA}  & \textbf{0.52} & 0.38 & 0.62 & \textbf{0.65} & 0.37 & 0.38 & \textbf{0.52} & 0.45 & 0.16 & 0.17 \\
\textbf{MSM}  & 0.43 & 0.36 & 0.50 & 0.48 & 0.20 & 0.22 & 0.51 & 0.48 & 0.05 & 0.09 \\

\bottomrule 
\end{tabular}

\label{table_correlation}
\end{table*}

\section{Experiments}

\subsection{Implementation Details}
\paragraph{Negative Summary Generation}
We randomly split the training set of CNN/DM dataset~\cite{nallapati-etal-2016-abstractive} in half and use half for training negative summarizer and the other half for generating negative summary after training. We use \textit{spaCy} for finding entities and noun phrases in both summaries and articles. We train \textit{bart-base}\footnote{https://huggingface.co/facebook/bart-base} for five epochs  to train MFMA, and use \textit{bart-base} model without fine-tuning for MF. We use \textit{t5-small}~\cite{raffel2020exploring}\footnote{https://huggingface.co/t5-small} for MSM, which shows better results than \textit{bart-base} for this task. We attach the further details in Appendix.

\paragraph{Training Classifier}
We train \textit{google/electra-base-discriminator}\footnote{https://huggingface.co/google/electra-base-discriminator} for five epochs with learning rate 2e-5, batch size of 96 using adam optimizer~\cite{kingma2015adam} with the dataset we generate using MF, MFMA and MSM. For DocNLI and FactCC, we get the original training dataset that each author release, and we train a model with the same setting as our method except for the training datasets for a fair comparison. We choose model using the balanced accuracy on validation set of FactCC~\cite{kryscinski-etal-2020-evaluating} which consists of 1k human annotated summaries.

\subsection{Benchmark Datasets}
For evaluating the performance of factual consistency checking system, it is necessary to compare the human judgments of the consistency for the summary with the system. And these human judgment exist in two forms, binary level(\textit{consistent}, \textit{inconsistent}) or numerical levels such as likert scale. In general, in the case of binary level data, performance is measured through accuracy with human judgments. For the case of numerical levels, correlation with human judgments is measured. In addition to using the results for the existing benchmark dataset in this way, we also report the accuracy by casting these numerical level datasets to the binary level dataset since we develop classifier based system. We report the results on the following datasets.

\paragraph{FC-Test}~\cite{kryscinski-etal-2020-evaluating} release a human-annotated factual consistency for the model generated summaries for CNN/DM Dataset in binary-level to test the performance of FactCC. There are 513 instances in this dataset.

\paragraph{XSumHall}~\cite{maynez2020faithfulness} study the types of hallucination in the generated summaries and collect the annotation on the errors in the 2K model generated summary for BBC XSum dataset~\cite{narayan-etal-2018-dont}. We use the datasets as binary level benchmark for XSum dataset as in~\cite{kryscinski-etal-2020-evaluating}.

\paragraph{SummEval}~\cite{fabbri2021summeval} collect the likert scale human judgments for the 1600 summaries generated from sixteen abstactive summarizer on CNN/DM testset. This dataset provides human judgments scores in terms of \textit{"coherence"
, "consistency", "fluency", and "relevance"} by three expert annotators in likert scale. We only use \textit{"consistency"} score of three annotators, for evaluating our proposed metric. For casting this score to binary level, we let the cases where at least one annotators give less than 5 points for \textit{"consistency"} as \textit{inconsistent}, otherwise \textit{consistent}.

\paragraph{QAGS-CNN/DM \& XSum}~\cite{wang-etal-2020-asking} release a human judgments for factual consistency on the model generated summaries for 235 summaries on CNN/DM testset and 239 summaries on XSum testset. Each summary is annotated by three annotators. We also cast the dataset to binary level by assigning \textit{inconsistent} if at least one annotators give \textit{inconsistent} label, otherwise \textit{consistent}.

\paragraph{FRANK-CNN/DM \& XSum}~\cite{pagnoni2021understanding} releases a benchmark dataset FRANK for summarization factual metrics which consists of 2246 summaries on the model generated summaries for 1250 summaries in CNN/DM and 996 summaries XSum. Three annotators evaluated factual consistency of the generated summaries in this dataset. We also convert this dataset to binary level as same as QAGS-CNN/DM and QAGS-XSum.

\subsection{Baseline Metrics}
We compare our methods with the following metrics. For all of the baseline metrics, we manually compute the score using the official repository which each author provided or reproducing the model for a fair comparison.

\paragraph{Entailment Based Metrics}
We adopt the model trained on MNLI~\cite{bowman2015large} and FEVER~\cite{thorne-etal-2018-fact} for factual consistency checking as in~\cite{kryscinski-etal-2020-evaluating}. FactCC~\cite{kryscinski-etal-2020-evaluating} and DocNLI~\cite{yin-etal-2021-docnli} are also entailment based models trained on synthetic dataset as in our work.

\paragraph{QA-Based Metrics}
QuestEval~\cite{scialom-etal-2021-questeval} uses the question generation and answering framework for evaluating the factual consistency of the summaries. QuestEval generates the question both the generated summaries and the source article, and then compare the answers of them with both summaries and the article to compute the factuality score of the summary.

\paragraph{N-gram Similarity Metrics}
BLEU~\cite{papineni-etal-2002-bleu}, ROUGE~\cite{lin-2004-rouge}, and METEOR~\cite{banerjee2005meteor} are widely used for evaluating the summaries. Among them, ROUGE-L, which uses F-measure based on the longest common subsequence between a candidate summary an the reference is the most widely used. 

\paragraph{Other Metrics}
BERTScore~\cite{zhang2020bertscore} utilizes cosine similarity of BERT~\cite{devlin2019bert} embeddings between the reference and the generated summary. CoCo~\cite{xie2021factual} computes the difference of likelihood of the summarizer between the summary with the original source and the summary with the masked source.

\subsection{Results}

\paragraph{Classification Accuracy}
Due to the imbalance in each dataset, we report the macro-F1 and class balanced accuracy in Table~\ref{table_accuracy_cnndm} and Table~\ref{table_accuracy_xsum}. We observe that macro-F1 score of our proposed methods MFMA outperforms baseline entailment metrics in five of seven benchmark datasets. MFMA shows better performances than other methods in especially for CNN/DM benchmarks, and shows similar performance to other baseline in XSum datasets. We explain that this is because we only use training set of CNN/DM to construct training set. On the other hand, DocNLI additionally uses the human annotated datasets from related tasks such as ANLI~\cite{nie2020adversarial} and SQuAD~\cite{rajpurkar2016squad} except for synthetic negative summaries. Another proposed method MSM also shows competitive performance for CNN/DM benchmarks, but relatively lower performance in XSum based benchmark datasets. We explain the performance gap between MSM and MFMA is due to the properties that directly generates summaries, resulting in many noisy samples that are relatively easy to be distinguished.


\paragraph{Correlation with Human Judgments}
To compare with general metrics that are not classification level, we also report the correlation with human judgments for five datasets in Table~\ref{table_correlation}.
We demonstrate that our proposed method has higher pearson correlation coefficient with human judgments in three of five benchmark datasets and competitive with the best results results in the spearman correlation coefficient. Especially, entailment based methods, which are relatively easy to compute, including our proposed methods show better results than QA-based QuestEval or likelihood based CoCo. Also, reference based methods such as ROUGE-L show very lower performance than other methods that do not require any references.

\begin{figure}[h]
\small
\centering
\includegraphics[width=1.0\columnwidth]{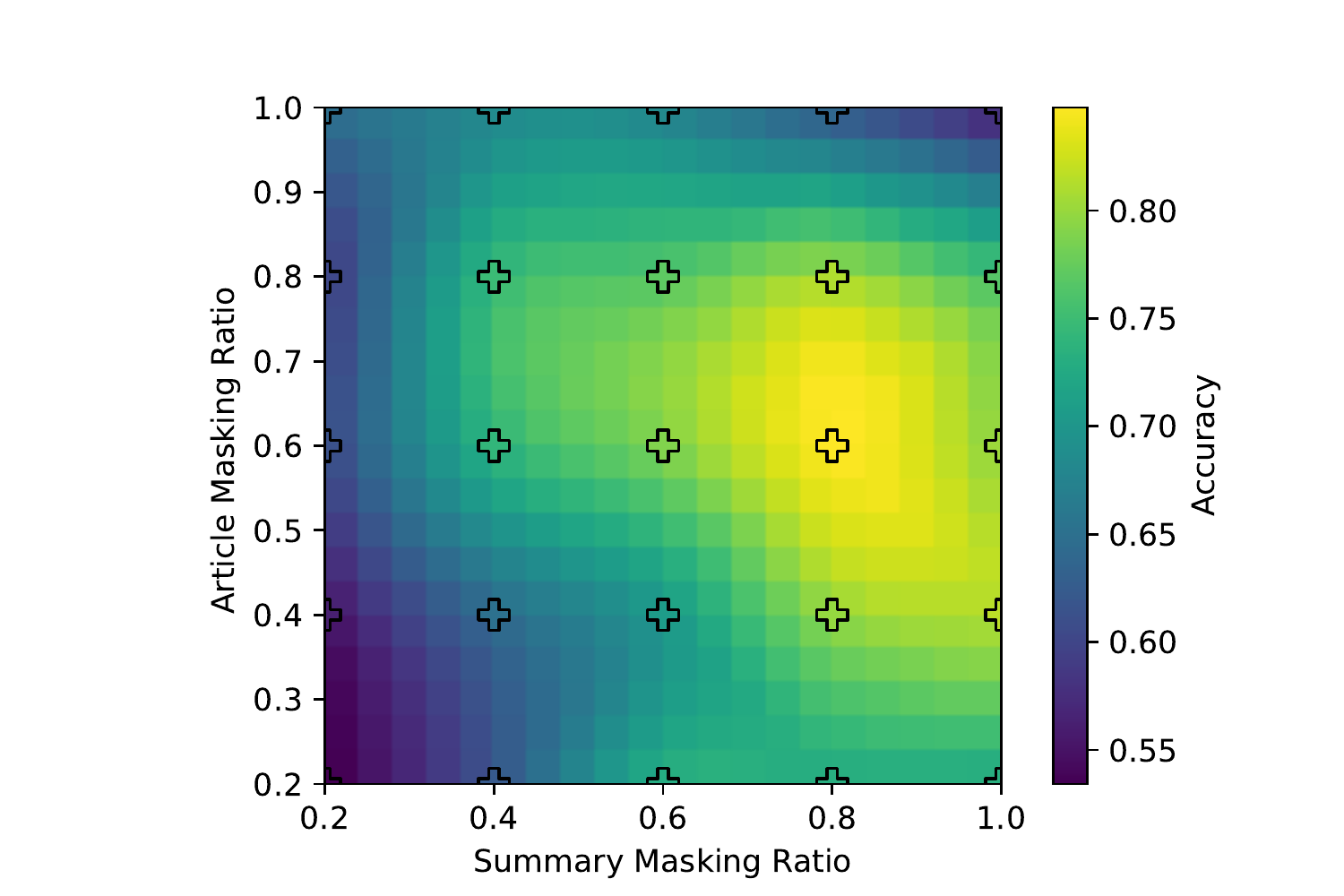}
\caption{
Validation Performance among Masked Ratio for Mask-and-Fill with Masked Article. We experiment with each of the five combinations of article mask ratio and summary mask ratio, and then plot the interpolated results.
}
\label{fig_2d_mask_ratio}
\end{figure}

\subsection{Analysis and Discussion} %

\paragraph{Performance among Masked Ratio} %
We analyze the effects of the mask ratio for both source article and summary in our proposed method MFMA and present results using the validation set in Figure~\ref{fig_2d_mask_ratio}. Through this experiment, we investigate the tradeoff in adjusting both the article masking ratio and summary masking ratio for generating negative summaries. As shown in Figure~\ref{fig_2d_mask_ratio}, we find that too high masking ratio decreases performance by sacrificing affinity. On the other hand, if the masking ratio is insufficient, the generated negative sample is often not really negative. In other words, too lower masking ratio leads to generate positive samples that are almost same as the original summary, and this degrades the performance of factual consistency checking model. Also, we can infer that there is an optimal masking ratio combination where the performance of factual consistency checking model is maximized.

\begin{figure}[t]
\scriptsize
\begin{framed}
\textbf{Article:} Tropical Storm Andrea formed in the Gulf of Mexico on Wednesday, marking the first storm of the 2013 Atlantic hurricane season. The storm was about 270 miles west-southwest of Tampa, Florida, and moving north at 6 mph, said the National Hurricane Center. It had maximum sustained winds of 40 mph. ,..., Thursday afternoon and move over southeastern Georgia and eastern South Carolina Thursday night and Friday. Little change in strength is expected over the next 48 hours.\\

\textbf{Reference Summary:} Storm moving north at faster rate (6 mph) A storm warning is issued for parts of the Florida coast . Little change in strength is forecast for the next 48 hours . Andrea the first storm of the 2013 Atlantic hurricane season . \\

\textbf{MF:} Storm moving north at 6 mph (11 kph )   A storm warning is issued for parts of the Carolinas.  Little change in strength is forecast for the Carolines.  Andrea the first storm of the season. \\

\textbf{MFMA $\gamma_A = 0.4$:} Andea moving north at faster rate (6 mph) A storm warning is issued for parts of the Florida coast. Little change in strength is forecast for the next 48 hours. Andea the first storm of the 2013 Atlantic hurricane season.
\\

\textbf{MFMA $\gamma_A = 0.6$:} Storm moving north at faster rate (6 mph) Tropical storm watch is issued for west coast of Florida. Little change in strength is forecast for next 48 hours. Storm the first storm of the 2013 Atlantic hurricane season. \\

\textbf{MFMA $\gamma_A = 0.8$:} Andrea moving north at 12 mph (6 mph) A tropical storm warning is issued for parts of the East Coast. A significant increase in strength is forecast for Thursday. Andrea the first storm of the Atlantic hurricane season. \\ 

\textbf{MFMA $\gamma_A = 1.0$:} Storm moving north at 6 mph (10 kph) A tropical storm watch is issued for the west coast of Florida. A hurricane watch in effect is forecast for the East Coast. The center of the storm forms. \\ 

\textbf{MSM $\gamma_A = 0.4$:} The storm is about 270 miles west-southwest of Fort Myers, Florida. NEW: The storm is in effect for the west coast of Florida. The storm is the first of the 2013 Atlantic hurricane season.

\end{framed}
\caption{
Generated negative summaries among various masking ratio in CNN/DM dataset. For MFMA and MF, we fix the summary masking $\gamma_S = 0.6$:
} 
\label{fig_ex_mask_ratio}
\end{figure}

\paragraph{Generated Samples among Masking Ratio}
We visualize the generated negative summaries through our proposed method MFMA and MSM using CNN/DM in Figure~\ref{fig_ex_mask_ratio}. We also visualize the example through MF, which simply fills in the mask without the article. We observe that if the article masking ratio $\gamma_A$ is too low, the generated summaries become almost similar to the original summary since there are enough information to fill the mask. However, if the $\gamma_A$ is too high, the generated examples are too far from the article, resulting in too negative summary similar to filling the mask without article. 

\begin{table}[h]
\small
\centering

\caption{Balanced accuracy of the human annotated factual consistency among masking unit. NP/Ent denotes \textit{noun phrases} and \textit{entities}.}

\resizebox{1\columnwidth}{!}{%
\begin{tabular}
{L{0.22\columnwidth}C{0.26\columnwidth}C{0.22\columnwidth}}
\toprule
\cmidrule{1-3}
           \textbf{Dataset} & \textbf{Avg-CNN/DM} & \textbf{Avg-XSum}\\
           
\midrule
\textbf{NP/Ent} & \textbf{73.9} & \textbf{60.9} \\
\textbf{Token} & 58.6 & 53.9 \\
\textbf{Sentence} & 53.5 & 53.4 \\
\bottomrule 
\end{tabular}
}
\label{table_masking_type}
\end{table}

\paragraph{Performance among Masking Unit}
We basically perform masking operation in the \textit{noun phrases} and \textit{entities} units for both summary and article. In order to see the effect of the masking unit, we also conduct an experiment on word level masking and sentence level masking, and present the classification level results in Table~\ref{table_masking_type}. We observe that \textit{noun phrases} level masking shows the best results following the work~\cite{goyal2021annotating} where many errors in summarization system are related to \textit{noun phrases} and \textit{entities}.

\begin{figure}[t]
\small
\centering
\includegraphics[width=1.0\columnwidth]{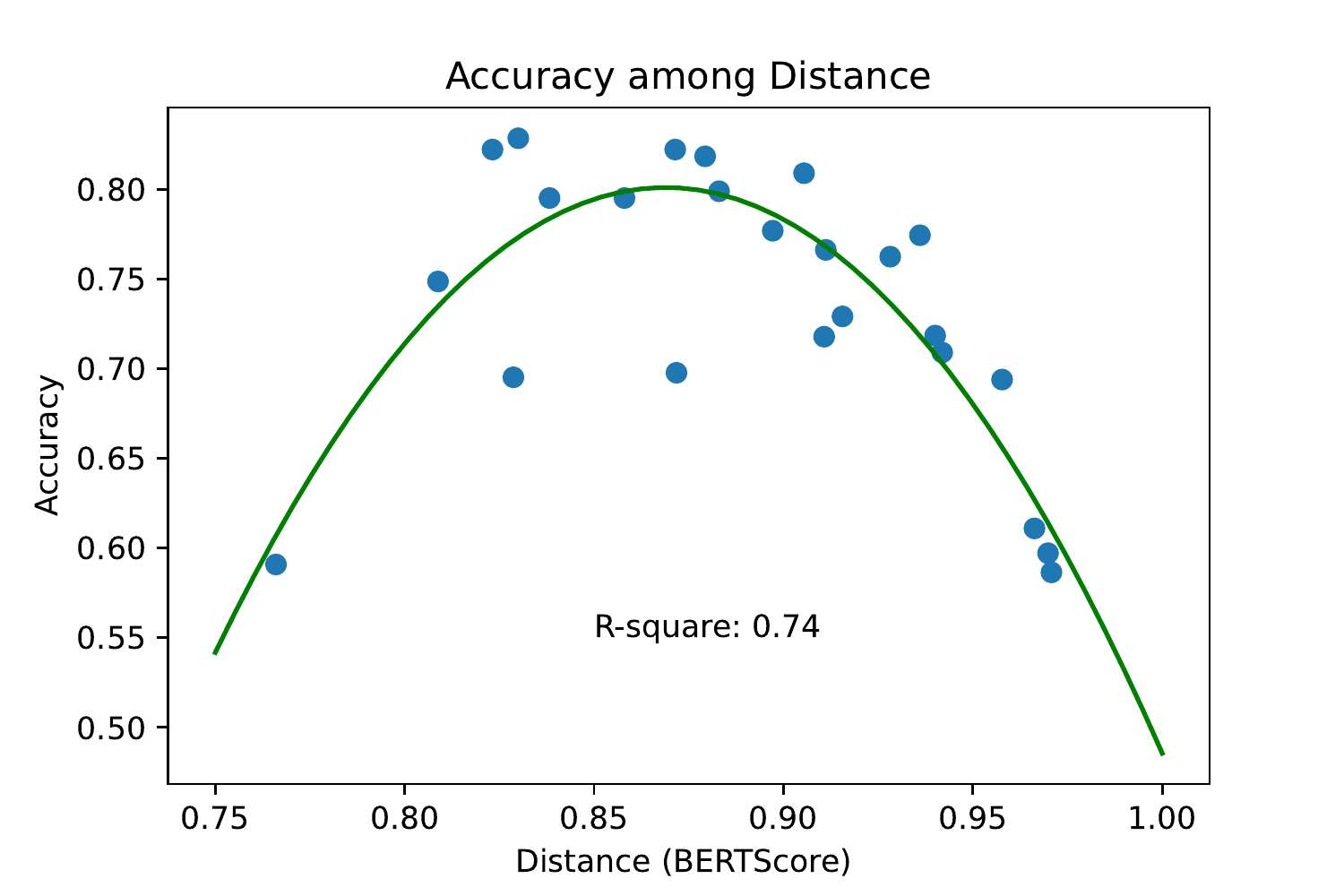}
\caption{
Validation Set Performance among BERTScore between the original reference summaries and the negative summaries we generate using the various combinations of article and summary masking ratios.  
}
\vspace{-5mm}
\label{fig_dist_acc}
\end{figure}
\begin{figure}[!t]
\small
\centering
\includegraphics[width=1.0\columnwidth]{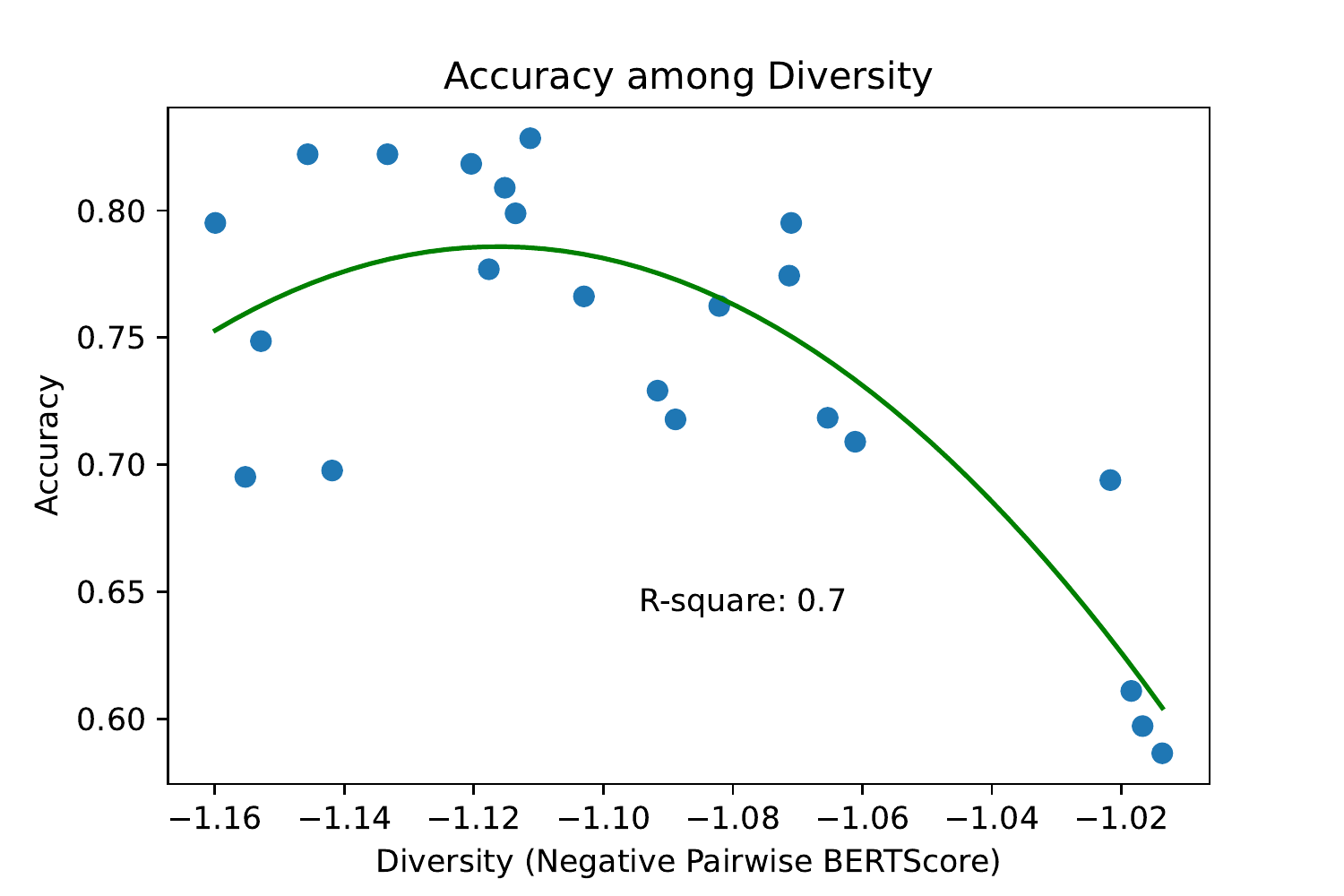}
\caption{
Validation Set Performance among diversity among various combinations of article masking ratio and summary masking ratio. Diversity is computed as negative of the pairwise BERTScore between four negative samples generated by each masking ratio.
}
\label{fig_dvs_acc}
\end{figure}

\paragraph{Distance from Original Reference Summary}
Using the results on various combinations of article masking ratio and summary masking ratio for MFMA as presented in Figure~\ref{fig_2d_mask_ratio}, we also investigate the relation between the average distance from the reference summary on each mask ratio combination and the performance. We compute BERTScore between original reference summary and the negative summary generated using the reference summary to get the distance. Interestingly, as shown in~Figure~\ref{fig_dist_acc}, we observe the distribution in which performance is maximized within the appropriate distance around 0.8 as the two-dimensional distribution with an $R^2$ of 0.74. This result shows how far the synthetic negative summaries must be from the reference summaries to help training the factual consistency checking model.

\begin{figure}[!t]
\scriptsize
\begin{framed}
\textbf{Article:} Nkaissery told reporters the university will be able to confirm Saturday if everyone has been accounted for. Thursday's attack by al-Shabaab militants killed 147 people, three security officers and two university security personnel. The attack left 104 people injured, including 19 who are in critical condition, Nkaissery said.,...,\\

\textbf{Candidate Summary:} 147 people, including 142 students, are in critical condition. \\

\textbf{Ground Truth:} \textit{INCONSISTENT}\\
\textbf{MFMA:} \textit{INCONSISTENT}\\ 
\textbf{MSM:} \textit{INCONSISTENT}\\ 
\textbf{DocNLI:} \textit{INCONSISTENT}\\
\textbf{FactCC:} \textit{CONSISTENT}\\

\textbf{Article:} Media playback is not supported on this device United remain 15 points clear at the top of the table with eight games left after a 1-0 win at Sunderland. "We are not concerned with what we have left behind us, we are only focusing on what is in front of us," said Ferguson. ",..., \\

\textbf{Candidate Summary:} Manchester United manager Sir Alex Ferguson says he is not concerned about his side's unbeaten start to the season as they attempt to win the Premier League title.\\

\textbf{Ground Truth:} \textit{CONSISTENT}\\
\textbf{MFMA:} \textit{INCONSISTENT}\\
\textbf{MSM:} \textit{INCONSISTENT}\\ 
\textbf{DocNLI:} \textit{INCONSISTENT}\\
\textbf{FactCC:} \textit{CONSISTENT}

\end{framed}
\caption{
Case study on entailment based models. First example comes from and FactCC-Test and second example comes from XSumHall.
}

\label{fig_case_study}
\end{figure}

\paragraph{Diversity among Masked Ratio}
Our proposed method can generate various samples depending on the location of the mask for the same summary-article pair with the fixed mask ratio.
Hence, we analyze the diversity of the generated negative summaries among the combinations of mask ratio for MFMA and present the result using validation set in Figure~\ref{fig_dvs_acc}. We define the diversity of each mask ratio combination as the negation of pairwise similarity score for each sample following~\cite{tevet-berant-2021-evaluating}. We sample four negative summaries using the given article for each method and then compute the pairwise similarity scores for all of the combinations. We also use BERTScore as a similarity measure. Similar to the distance, we observe that diversity has also similar to a two-dimensional form with an $R^2$ of 0.7, in which the accuracy is maximized at an appropriate point.

\paragraph{Case Study}
To understand the pros and cons of our proposed factual consistency checking system, we conduct a case study and illustrate the representative success and failure cases in Figure~\ref{fig_case_study}. We observe that our system is good at judging the facts themselves in the summary like the first example, but still not perfect in examples that require high-level reasoning like the second example. We expect the system can be improved by adopting MFMA and MSM to the datasets that have more abstractive summaries which require more reasoning to check the factual consistency.

\section{Conclusion}
In this paper, we proposed an effective generation method of factually inconsistent summaries, called MFMA. In this method, some proportion of the source text and corresponding reference summaries is hidden, then a summarization model generates plausible but factually inconsistent summaries by inferring the masked contents.
Experiments on seven benchmark datasets demonstrate that factual consistency classifiers trained using our method generally outperform existing models and show a competitive correlation with human judgment.

\section*{Ethical Considerations}
Our approach creates a synthetic dataset using a public dataset to train a factual consistency checking model. Therefore, in the process of generating such samples, ethically problematic datasets can be generated due to the bias of the pre-trained models, similar to other text generation tasks. For this reason, once the training process is completed, we remove the generated sample. And, we will not release the synthetic dataset itself, and will release only the trained factual consistency checking model.

\section*{Acknowledgements}
K. Jung is with ASRI, Seoul National University, Korea. This research was supported by SNU-NAVER Hyperscale AI Center.
\bibliography{anthology,acl}
\bibliographystyle{acl_natbib}

\clearpage

\appendix
\section{Experimental Details}

\subsection{Reproducibility Checklist}

\paragraph{Source Code}
We attach the source in the submission and we will release the pre-trained factual consistency checking model.

\paragraph{Computing Infrastructure}
We use Intel(R) Xeon(R) Silver 4210R CPU (2.40 GHz) with NVIDIA RTX A5000 24GB for the experiments. The software environments are Python 3.8.8 and PyTorch 1.10.1.

\paragraph{Dataset Statistics}
We use the training of CNN/DM dataset that consists of 287113 examples. We divide it in half randomly and use one for MSM or MFMA training and the other for generating negative summaries. Then, we merge the generated article-negative summaries pairs and the article-positive summaries we used for training MFMA and MSM to construct the training set for factual consistency checking model.

\paragraph{Average runtime for each approach}
For training MFMA and MSM, it takes 10 hours to train the whole model. And it takes 3 hours to generate whole negative summaries that is to be used for training factual consistency checking. For training factual consistency checking model, it takes 7 hours using a single GPU.

\paragraph{Hyperparameters}
We train five epochs for MFMA and MSM using \textit{bart-base} for MFMA and \textit{t5-small} for MSM respectively. We train the model with batch size of 48, max input sequence size of 1024, and max target sequence size of 140. We conduct experiment with various article masking $\gamma_A$ ratio-summary masking ratio $\gamma_S$ combinations, at 0.2 intervals from (0.2, 0.2) to (1.0, 1.0). For the case of training classifier, we train \textit{google/electra-base-discriminator} for five epochs with learning rate 2e-5 and batch size of 96. We choose the best parameters using the validation set provided by the~\cite{kryscinski-etal-2020-evaluating}. The best mask ratio combination is $\gamma_A$ = 0.6 and $\gamma_S$ = 0.8.

\paragraph{Number of Model Parameters} 
The number of parameters for negative summary generation model is 139M for MFMA, is 0.6M (t5-small) and the factual consistency classifier is 109M.

\subsection{Computing Baseline Metrics}
Even with the same dataset, the results may be different due to some factors such as type of tokenizer or case, so we calculate baseline ourselves as follows.
For n-gram similarity metrics BLEU-4, ROUGE-L and METEOR, we compute the scores using the package \textit{language evaluation}\footnote{https://github.com/bckim92/language-evaluation} which is based on COCOeval\footnote{https://github.com/tylin/coco-caption}.
For BERTScore\footnote{https://github.com/Tiiiger/bert_score}, QuestEval\footnote{https://github.com/ThomasScialom/QuestEval} and CoCO\footnote{https://github.com/xieyxclack/factual_coco}, we use the official repository with the default setting. For MNLI, we use \textit{roberta-large-mnli}\footnote{https://huggingface.co/roberta-large-mnli} and use {tals/albert-base-vitaminc-fever}\footnote{https://huggingface.co/tals/albert-base-vitaminc-fever} for FEVER.


\subsection{Significance Test}
We adopt standard way to test the significance of the correlation coefficient for all of the reported related correlation coefficients in Table~\ref{table_correlation}. We compute the p-value for each coefficient with a t-test that uses a null hypothesis, which is an absence of association.

\end{document}